%% file: arxiv.tex
\documentclass[english]{article}
\input{config/packages.sty}
\input{config/commands.sty}

\title{Markov Chain-based Optimization Time Analysis of Bivalent Ant Colony Optimization for Sorting and LeadingOnes
\footnote{This is the preprint of a full-paper accepted for presentation at the Genetic and Evolutionary Computation Conference (GECCO), 2024.}}
\author{
	\orcidauthor{Matthias Kerga\ss ner}{0000-0003-2830-2679}\footnote{Corresponding author}\\
	\email{matthias.kergassner@fau.de}
	\and
	\orcidauthor{Oliver Keszocze}{0000-0003-2033-6153}\\
	\email{oliver.keszoecze@fau.de}
	\and
	\orcidauthor{Rolf Wanka}{0000-0002-5729-3727}\\
	\email{rolf.wanka@fau.de}
}

\date{Department of Computer Science, Friedrich-Alexander-Universit\"at Erlangen-N\"urnberg, Germany}

\begin{document}
\maketitle
\begin{abstract}
	\input{sections/abstract.tex}

\end{abstract}
\input{sections/introduction.tex}
\input{sections/baco.tex}
\input{sections/runtime_analysis/runtime_analysis.tex}
\input{sections/experiments/experiments.tex}
\input{sections/conclusion.tex}

\appendix
\input{sections/appendix/appendix.tex}

\newpage
\bibliographystyle{acm}
\bibliography{literature}
\end{document}

%% file: sections/abstract.tex
So far, only few bounds on the runtime behavior of Ant Colony Optimization (ACO) have been reported.
To alleviate this situation, we investigate the ACO variant we call Bivalent ACO (\baco) that uses exactly two pheromone values.
We provide and successfully apply
a new Markov chain-based approach to calculate the expected optimization time, i.\,e., the expected number of iterations until the algorithm terminates.
This approach allows to derive exact formul\ae{} for the expected optimization time for the problems \sorting and \leadingones.
It turns out that the ratio of the two pheromone values significantly governs the runtime behavior of \baco.
To the best of our knowledge, for the first time, we can present tight bounds for \sorting ($\Theta(n^3)$) with a specifically chosen objective function  and prove the missing lower bound $\Omega(n^2)$ for \leadingones which, thus, is tightly bounded by $\Theta(n^2)$.
We show that despite we have
a drastically simplified ant algorithm with respect to the influence of the pheromones on the solving process, known bounds on the expected optimization time for the problems \onemax ($\mathcal{O}(n \log n)$) and \leadingones ($\mathcal{O}(n^2)$) can be re-produced as a by-product of our approach.
Experiments validate our theoretical findings.

%% file: sections/introduction.tex
\section{Introduction}
Ant algorithms are nature-inspired meta-heuristics that imitate colonies of ants building up an efficient path between their nest and a source of food.
They quickly gained a lot of attention in the scientific community due to their ability to be adapted to a large variety of problems in particular combinatorial optimization problems.
Initially the algorithms Ant System, Ant Colony System and Ant Colony Optimization (ACO) were introduced \cite{dorigo1992,dorigo1999a,gambardella1996} and further developed to the Min-Max-Ant System \cite{stutzle1997max,STUTZLE2000889}.
Ant algorithms have been applied to a variety of problems, e.\,g., their intuitive vocation of Shortest Paths~\cite{SUDHOLT2012165} but also the Traveling Salesperson Problem \cite{782657,stutzle1997max}, Minimum Spanning Tree \cite{NEUMANN20102406}, Minimum Cut \cite{koetzing2010}, Minimum Weight Vertex Cover \cite{Shyu2004} and Examination Time Tabling \cite{abounacer:hal-01286691}.
Besides those combinatorial optimization problems, ACO has been studied for the class of pseudo-boolean function, e.\,g., \onemax, \leadingones and \binaryvalue \cite{GUTJAHR20082711,neumann2006,doerr2007a}.
Although these problems are not considered real-world applications, their analysis yields
many insights into and understanding of ACO.
Furthermore they are typical benchmarks that allow comparisons of the optimization time to other meta-heuristics such as Evolutionary Algorithms (EA) and Particle Swarm Optimization (PSO).
Optimization time, i.\,e., the number of iterations of an algorithm, is a fundamental matter in the context of algorithmic fields.
Scharnow et al.~\cite{ScharnowTW04} analyzed the expected optimization time of an evolutionary algorithm for \sorting and single-source-shortest-paths (SSSP).
Doerr et al.~\cite{4424704} improved that result for SSSP further and achieved tight bounds.
M\"u{}hlenthaler et al.~\cite{Muehlenthaler2021} proved bounds on the expected optimization time of PSO applied to \sorting and \onemax.
Neumann and Witt \cite{neumann2009runtime} proved that the expected optimization time of ACO for any linear pseudo-boolean function is bounded by $\bigO(n\log n)$.
A wide and up-to-date overview on the topic of evolutionary computation and recent developments is given in \cite{doerr2020}.

The aim of this work is to contribute to the successful analyses of ACO by providing an approach that yields novel tight bounds on the expected optimization time for \sorting ($\Theta(n^3)$).
Furthermore we demonstrate that our approach produces tight bounds on the expected optimization time for \leadingones ($\Theta(n^2)$).
So far, to the best of our knowledge, the lower bound for \leadingones was an open question \cite{DOERR20111629}.
Overall we provide a powerful tool that will be deployed to investigate further optimization time bounds of ACO.

%% file: sections/baco.tex
\section{Bivalent Ant Colony Optimization}
\label{sec:baco}
Ant Colony Optimization is an algorithmic approach to solve black-box optimization problems.
The input of an ACO algorithm is a construction graph $G$ and an objective function $f$.
The latter maps any path the ant\footnote{we use a single ant; see discussion below} walks on $G$ to a real number, representing the quality of the solution derived from this path.
Depending on the pheromone strategy, pheromones are spread along the edges of that path to influence its probability to be incorporated into the next path.
A high pheromone value increases the probability of an edge to be in the path of the subsequent iteration.
The counterpart is the evaporation of pheromones over time.
Overall, pheromones serve as memory where good solutions are enforced and weak ones decay over time.
Thus finding the right balance between the spreading and evaporation of pheromones is essential.

\begin{algorithm}[t]
	\caption{\baco}
	\label{alg:baco}

	\SetKwInOut{Input}{Input}
	\SetKwInOut{Parameter}{Parameter}
	\SetKwInOut{Output}{Output}
	\DontPrintSemicolon

	\Input{Construction graph $G$, objective function $f$}
	\Parameter{Pheromone levels $\taumin$ and $\taumax$\newline random walk policy $\mathcal{P}$}
	\Output{Best path $\pi^*$ found}

	Initialize all pheromones with $\taumin$\;
	The ant constructs an initial best path $\pi^*$ by applying $\mathcal{P}$\;
	\While{\textup{termination criterion not fulfilled}}
	{
		The ant constructs a path $\pi$ on $G$ by applying $\mathcal{P}$\;
		\If{$f(\pi)>f(\pi^*)$}
		{
			$\pi^*\leftarrow\pi$\;
			Update pheromones\;
		}
	}
	\Return $\pi^*$\;
\end{algorithm}
We
present the Bivalent Ant Colony Optimization algorithm (\baco) in the following (see~Algorithm~\ref{alg:baco}).
The pheromones in \baco are bivalent and can only take the values $\taumin$ and $\taumax$.
Let $G$ be the construction graph and $f$ the objective function as described above.
Initially, every egde of $G$ has the pheromone level $\taumin$.
\baco uses a single ant that walks across $G$ according to the random walk policy $\mathcal{P}$.
This policy prescribes on which node the ant starts, when the ant should stop and can, e.\,g., prevent the ant from visiting a node more than once (cf.~tabu list).
Such random walk rules highly depend on the problem and the adaptation of ACO to it.
The first random walk of the ant yields the initial best path $\pi^*$.
During every iteration the ant builds up a path $\pi$ on $G$ that is evaluated via the objective function $f$.
If $\pi$ is better than $\pi^*$, then $\pi^*$ is replaced by $\pi$ and pheromones are updated.
Therefore every edge that is used in path $\pi^*$ obtains the pheromone value $\taumax$ while all the remaining edges are assigned the $\taumin$ pheromone value.
In this work, the termination criterion is chosen such that \baco iterates until an optimum solution has been found.
As all problems investigated have a known optimal solution, this is a reasonable criterion.

It is common in the analysis of ant algorithms to use a single ant only~\cite{neumann2009runtime,NEUMANN20102406}.
Also in this work, having more than one ant would not yield any advantage due to the following reasoning.
If $k$ ants walk across $G$ they produce $k$ paths per iteration.
In case there is a new best path among these, the ants will not benefit from the corresponding spread of pheromones until the next iteration.
This results in an increased number of objective function evaluations.
If actual multi-threading is used, the runtime of the program will improve when $k$ scales with the number of CPU cores.
However, this does not affect the computational complexity that is calculated in this work since we investigate the number of objective function evaluations.
Thus in the following the number of evaluations equals the number of iterations.

Another characteristic of \baco
is the bivalent pheromones.
This serves two purposes at the same time.
Bounding pheromones limits the evaporation which fundamentally improves the runtime of the algorithm~\cite{stutzle1997max,STUTZLE2000889,DORIGO2005243,NEUMANN20102406}.
If pheromones were unbounded, the pheromone levels on many edges decay quickly such that they will probably never be chosen by the ant.
Vice versa, we precisely calculate such bounds to ensure that the ant can pick any path with a reasonable probability even if numerous iterations have already passed.
In fact, the crucial parameter is the pheromone ratio $t=\nicefrac{\taumin}{\taumax}$ and not the actual values of $\taumin$ and $\taumax$.
Doerr et al.~\cite{DOERR20111629} show that the right balance of pheromone bounds is not only a tool to fine-tune ACO algorithms, but that ACO is not robust with respect to the ratio $t$.
It also turns out in this paper that the parameter $t$ is crucial for the runtime behavior of BACO.
Therefore, $t$ needs to be determined carefully.
The second purpose of bivalent pheromones is the significant simplification of evaporation.
As described in the pheromone update mechanism above, pheromones evaporate from $\taumax$ to $\taumin$ immediately, if the corresponding edge is not enforced in the update.

In the subsequent analysis, we analyze \baco precisely and demonstrate that, despite its simplifications, it is competitive.
We will present the decisive role of the pheromone ratio $t$ on the expected optimization time and use this insight to identify optimal pheromone ratios.

%% file: sections/runtime_analysis/runtime_analysis.tex
\section{Markov Chain-based Optimization Time Analysis of BACO}
\label{sec:runtime_analysis}
We quantify the progress of \baco using the objective function $f$.
Let $m$ be the number of elements of the codomain of $f$ and let the Markov states $0$ through $m-1$ be identified with the smallest up to the largest function value respectively.
Then the final state $m-1$ corresponds to the optimum of $f$.
The current Markov state depends on the objective value of the current best solution.
Let $p$ be the probability distribution vector of the first Markov state, i.\,e., $p(i)$ is the probability that the initial Markov state is $i$.
Denote the probability to reach state $j$ from $i$ within one iteration of the algorithm by $\Pr[i\rightarrow j]$.
Then the $(m\times m)$-Markov matrix $M$ is defined elementwise by $M(i,j)=\Pr[i\rightarrow j]$.
Set $\Pr[m-1\rightarrow j]=0$ for any value $j$ since the algorithm terminates as soon as state $m-1$ is reached.
This facilitates the subsequent calculations.
Furthermore $M$ is an upper triangular matrix, since \baco discards candidate solutions that are worse than the current best, i.\,e., $\Pr[i\rightarrow j]=0$ for $j<i$.

The following theorem yields a formula to calculate the expected optimization time of \baco exactly.
To make this work the Markov process has to have another property, namely for every state $i$ the probability to leave it has to be positive.
In other words, the probability $\Pr[i\rightarrow i]$ to stay in a non-optimal state $i$ must not equal $1$.
Otherwise the process could stagnate prematurely in such a sub-optimal state.
\begin{theorem}
	\label{thm:T_with_inverse}
	Let $p$ be the probability distribution vector of the initial Markov state and $M$ the Markov matrix.
	The expected optimization time $T(n,t)$ of \baco with the pheromone ratio $t$ is given by
	\begin{align}
		T(n,t)
		=p\cdot(\eye-M)^{-1}\cdot(\eye-M)^{-1}\cdot M\cdot(0,\dots,0,1)^\top, \label{eq:T_with_inverse}
	\end{align}
	where $\eye$ denotes the unit matrix and the vector $(0,\dots,0,1)^\top$ consists of zeros and a one as the last entry, both of fitting dimension.
	\begin{proof}
		The probability distribution vector $p$ of the initial Markov state multiplied with the Markov matrix $M$ raised to the power $i$, $i\ge 0$ yields a vector where the $j$th entry equals the probability to be in state $j$ after $i$ iterations.
		So $p\cdot M^i\cdot (0,\dots,0,1)^\top$ equals the chance to be in the final state $m-1$ after exactly $i$ iterations.
		With that the expected optimization time is calculated as follows:
		\begin{align*}
			T(n,t)
			&=\sum_{i=0}^{\infty} \left(i\cdot p\cdot M^i\cdot(0,\dots,0,1)^\top\right)\\
			&=p\cdot\left(\sum_{i=0}^{\infty} \left(i\cdot M^i\right)\right)\cdot(0,\dots,0,1)^\top\\
			&=p\cdot(\eye -M)^{-1}\cdot(\eye -M)^{-1}\cdot M\cdot(0,\dots,0,1)^\top
		\end{align*}
		Note that $\eye -M$ is invertible since its eigenvalues are non-zero due to the following reasoning.
		$M$ is an upper triangular matrix and its diagonal entries are in the right-open interval $[0,1[$ as discussed above.
		Thus $\eye -M$ is also upper triangular and its diagonal entries are strictly positive and in the left-open interval $]0,1]$.
		Consequently the eigenvalues of $\eye -M$ are non-zero, so the inverse exists.
		A detailed reasoning to the limit of the infinite sum of powers of $M$ can be found in Appendix~\ref{sec:infinite_sum_of_matrix_powers}.
	\end{proof}
\end{theorem}
\input{sections/runtime_analysis/T_without_inverse.tex}
In the next sections, two problems are investigated where this formula applies, namely \leadingones and \sorting.
The results are discussed in Section~\ref{sec:discussion_of_results}.
\input{sections/runtime_analysis/leadingones.tex}

We also investigated \baco applied to the pseudo-boolean \onemax function.
From that we can confirm the known upper bound $\mathcal{O}(n\cdot\log n)$ in \cite{GUTJAHR20082711}.
However, since this proof does not contribute new insights to the topic we omit it in this work.
\input{sections/runtime_analysis/sorting.tex}

\subsection{Discussion of the Results}
\label{sec:discussion_of_results}
In this section the results of the preceding analyses of \baco are discussed.
Table~\ref{tab:comparison_runtime_aco_ea_pso} presents the expected optimization time bounds for \baco, the \oneoneea from \cite{DROSTE200251} and \onepso from \cite{Muehlenthaler2021} applied to \leadingones and \sorting (no results are available for \onepso applied to \leadingones).
\input{sections/runtime_analysis/table_runtime_comparison.tex}

In Section~\ref{sec:leadingones} we provide an exact formula for the expected optimization time of \baco for \leadingones (Eq.~\eqref{eq:expected_leadingones}).
For the pheromone ratio $t=\nicefrac{c}{n^s}$ with $s\ge 1$ we prove that this formula is tightly bounded by $\Theta(n^{s+1})$.
If $s$ is in the interval $[0,1[$ then this bound becomes superpolynomial.
Thus, when $s$ is chosen smaller than $1$ then there is a phase transition from polynomial to superpolynomial complexity.
Overall $t=\nicefrac{c}{n}$ is optimal with respect to the expected optimization time $\Theta(n^2)$.
Note that the upper bound $\bigO(n^2)$ has already been shown in \cite{doerr2007a} but that so far no lower bound was known.
Additionally, we prove that for any $t\in\bigO(1)$ the expected optimization time is tightly bounded by $\Theta\big(\frac{1}{t^2}\cdot((1+t)^n-1)\big)$.

Next we utilize our approach to obtain an exact formula for the expected optimization time $T(n,t)$ of \baco for \sorting (Eq.~\eqref{eq:expected_sorting}).
We prove that for any pheromone ratio $t\in\bigO\big(\frac{1}{n^2}\big)$ the expected optimization time is tightly bounded by $\Theta\big(\frac{n}{t}\big)$.
In particular, if \mbox{$t=\nicefrac{c}{n^s}$} and $s\ge 2$, then $T(n,t)$ is tightly bounded by $\Theta(n^{s+1})$.
Together with the lower bound $\Omega(n^3)$ for any (positive) pheromone ratio $t=t(n)$ it follows that $t=\nicefrac{c}{n^2}$ is optimal with respect to the expected optimization time which, consequently, is bounded by $\Theta(n^3)$.
For $t=\nicefrac{c}{n^s}$ and $s<2$ we prove the upper bound $\bigO\big(n^{s+1}\cdot \mathconst{e}^{c\cdot n^{2-s}}\big)$.
We assume that for $s=2$ the analog phase transition as for \leadingones from polynomial to superpolynomial complexity occurs.
However, this remains an open question.
If $t$ is set to~$1$, then $\taumin$ equals $\taumax$ and thus no information is gained from the pheromones.
This leads to a blind search and an expected number of iterations bounded by $\Theta(n!)$.

Note that standard application of Markov's inequality leads to the observation that the probability that \baco is successful in $\alpha\cdot T(n,t)$ optimization steps is at least $1-\nicefrac 1 {\alpha}$.

Overall \baco has the
same expected optimization time for \leadingones as the \oneoneea, assuming both algorithms use their corresponding optimal parameter settings.
In the domain of \sorting however, \baco is less efficient than the \oneoneea and \onepso by a factor of $\nicefrac{n}{\log n}$.
Further research might lead to a competitive optimization time of \baco for \sorting.

%% file: sections/runtime_analysis/T_without_inverse.tex
The computation of the inverse in Theorem~\ref{thm:T_with_inverse} should be avoided as it quickly becomes
time consuming and numerically unstable.
Therefore Eq.~\eqref{eq:T_with_inverse} is substituted by an explicit formula in the following.
For this, the Markov matrix $M$ needs to fulfill two properties:
\begin{itemize}
	\item
		Each row of $M$ sums up to $1$ except for the last row.
		The sum of the last row does not play a role in the subsequent proof.
	\item
		$\forall i,j$, $0\leq i<j\leq m-1$, $i+1<j$:
		\begin{align}
			\frac{M(i,j)}{M(i+1,j)}=\phi_i\label{eq:row_property_M}
		\end{align}
		This means that these ratios of values of $M$ only depend on the row index $i$ and not on the column.
\end{itemize}
Let $A=\eye -M$ for the remainder of this work.
Note that the same values $\phi_i$ apply to $A$ by definition.
Consider the following Theorem.
\begin{theorem}
	The inverse matrix of $A=\eye -M$ takes the form
	\begin{align*}
		A^{-1}(i,j)=
		\begin{cases}
			\frac{1}{A(i,i)}	& \text{if $i=j$}\\
			\delta_j			& \text{if $i<j$}\\
			0					& \text{else}
		\end{cases},
	\end{align*}
	where $\delta_j$, $1\le j\le m-1$ is defined as
	\begin{align}
		\delta_j
		=-\frac{A(j-1,j)}{A(j-1,j-1)\cdot A(j,j)}\label{eq:delta_j}.
	\end{align}
	\begin{proof}
		Since $M$ is an upper triangular matrix, $A=\eye -M$ has the same property.
		By prerequisite the diagonal entries of $M$ are not equal to $1$.
		Thus the diagonal of $A$ is non-zero and the diagonal of $A^{-1}$ consists of the corresponding inverse values.
		
		What is left to prove is $A^{-1}(i,j)=\delta_j$ for all $i<j$.
		First prove by mathematical induction that $A^{-1}(i,m-1)$ equals $1$ for all $i$.
		$A^{-1}(m-1,m-1)=1$ follows directly from the definition of $M$.
		Now assume $A^{-1}(k,m-1)=1$ holds true for all $k>i$ and conclude the claim for $A^{-1}(i,m-1)$.
		\begin{align*}
			0
			\overset{!}{=}{}&
				\left(A\cdot A^{-1}\right)(i,m-1)
			=
				\sum_{k=i}^{m-1} A(i,k)\cdot A^{-1}(k,m-1)\\
			={}&
				A(i,i)\cdot A^{-1}(i,m-1)+\sum_{k=i+1}^{m-1} A(i,k)\cdot\underbrace{A^{-1}(k,m-1)}_{=1}
		\end{align*}
		The sum of each row of matrix $M$ except the last equals $1$.
		Thus all rows of $A$ except for the last one sum up to $0$.
		With that the former calculation is continued.
		\begin{align*}
			0
			\overset{!}{=}{}&
			A(i,i)\cdot A^{-1}(i,m-1)+\overbrace{\sum_{k=i+1}^{m-1} A(i,k)}^{\overset{\text{every row sum is $0$}}{=}-A(i,i)}\\
			={}&
			A(i,i)\cdot A^{-1}(i,m-1)-A(i,i)
		\end{align*}
		This implies that $A^{-1}(i,m-1)=1$ holds.
		By induction the last column of $A^{-1}$ consists of ones only.
		Next, prove the off-diagonal entries.
		\begin{align*}
			0
			\overset{!}{=}{}&
				\left(A\cdot A^{-1}\right)(j-1,j)\\
			={}&
				A(j-1,j-1)\cdot A^{-1}(j-1,j)+A(j-1,j)\cdot A^{-1}(j,j)\\
			={}& A(j-1,j-1)\cdot A^{-1}(j-1,j)+\frac{A(j-1,j)}{A(j,j)}
		\end{align*}
		Now solve this equation for $A^{-1}(j-1,j)$ to obtain the claim
		\begin{align*}
			A^{-1}(j-1,j)
			=-\frac{A(j-1,j)}{A(j-1,j-1)A(j,j)}
			=\delta_j.
		\end{align*}
		Finally, apply Eq.~\eqref{eq:row_property_M} to prove the rest.
		Let $i,j$, $0\leq i,j\leq m-1$ and $i+1<j$.
		\begin{align*}
			0
			\overset{!}{=}{}&
				\left(A\cdot A^{-1}\right)(i,j)
			=
				\sum_{k=i}^{j} A(i,k)\cdot A^{-1}(k,j)
			=
				A(i,i)\cdot A^{-1}(i,j)\\
			&
				+A(i,i+1)\cdot A^{-1}(i+1,j)
				+\sum_{k=i+2}^{j} \underbrace{A(i,k)}_{\mathrlap{\overset{\eqref{eq:row_property_M}}{=}\phi_i\cdot A(i+1,k)}}{}\cdot A^{-1}(k,j)\\
			={}&
				A(i,i)\cdot A^{-1}(i,j)+A(i,i+1)\cdot A^{-1}(i+1,j)\\
			&
				+\phi_i\cdot\underbrace{\sum_{k=i+2}^{j} A(i+1,k)\cdot A^{-1}(k,j)}_{\mathclap{\overset{\left(A\cdot A^{-1}\right)(i+1,j)=0}{=}-A(i+1,i+1)\cdot A^{-1}(i+1,j)}}\\
			={}&
				A(i,i)\cdot A^{-1}(i,j)\\
			&+
				A^{-1}(i+1,j)\cdot\left(A(i,i+1)-\phi_i\cdot A(i+1,i+1)\right)
		\end{align*}
		Rearrange the equation to the following form:
		\begin{align*}
			\frac{A^{-1}(i,j)}{A^{-1}(i+1,j)}
			=\frac{1}{A(i,i)}\cdot\left(\phi_i\cdot A(i+1,i+1)-A(i,i+1)\right)
		\end{align*}
		The fraction on the left side of the equation is independent of the column index $j$ since $j$ does not appear on the right side.
		Since the last column of $A^{-1}$ consists of ones only, the fraction itself must be equal to $1$.
		Thus the fraction equals $1$ for all the other columns respectively.
		Overall this implies that in column $j$ of $A^{-1}$ all entries above the diagonal take the same value $\delta_j$.
	\end{proof}
\end{theorem}
Now the exact values of $A^{-1}$ are known and Eq.~\eqref{eq:T_with_inverse} can be simplified as follows.
\begin{theorem}
	\label{thm:T_without_inverse}
	Let $p$ and $M$ be defined as in Theorem~\ref{thm:T_with_inverse}.
	If each row of $M$ but the last sums up to $1$ and the property in Eq.~\eqref{eq:row_property_M} is fulfilled, then the expected optimization time $T(n,t)$ of \baco can be calculated explicitly.
	\begin{align}
		T(n,t)
		=\sum_{i=0}^{m-2} \left(p(i)\cdot\left(\frac{1}{A(i,i)}+\sum_{j=i+1}^{m-2}\delta_j\right)\right)
	\end{align}
	\begin{proof}
		Start with Eq.~\eqref{eq:T_with_inverse} and use the identity $M=\eye -A$.
		Then exploit the fact that each value in the last column of $A^{-1}$ equals $1$.
		\begin{align*}
			&
				T(n,t)
			=
				p\cdot A^{-1}\cdot A^{-1}\cdot (\eye -A)\cdot (0,\dots,0,1)^\top\\
			={}&
				p\cdot A^{-1}\cdot\underbrace{(A^{-1}-\eye)\cdot (0,\dots,0,1)^\top}_{=\text{ last column of }A^{-1}-\eye}
			=
				p\cdot A^{-1}\cdot (1,\dots,1,0)^\top
		\end{align*}
		Write out the multiplications to obtain the claim.
	\end{proof}
\end{theorem}

%% file: sections/runtime_analysis/leadingones.tex
\subsection{\textsc{LeadingOnes}}
\label{sec:leadingones}
The \leadingones function counts the number of consecutive ones from the beginning of a bit string of length $n$.
Thus the objective value is a number from $0$ through $n$.

For pseudo-boolean objective functions such as \leadingones, a directed construction graph $G$ called \emph{chain}  is used~\cite{GUTJAHR20082711,DOERR20111629}.
The starting node is connected to two nodes which denote the first bit of $x$ to be $0$ or $1$ respectively.
These two nodes are connected to a merge-node.
From here the same structure repeats for the second bit of $x$ etc.
This way an ant walking across $G$ builds up a path that is identified with a bit string.

We will see that the \leadingones problem fulfills the requirements for Theorem~\ref{thm:T_without_inverse} allowing us to calculate the expected optimization time of \baco.

We first calculate the probability distribution of the initial Markov state and the Markov matrix for \leadingones and then derive the exact formula for the expected optimization time as well as bounds on it.
\begin{observation}\label{obs:p_leadingones}
	Let $m=n+1$ be the number of Markov states.
	The first path across $G$ constructed by the ant yields a uniformly at random chosen candidate solution for \leadingones.
	The probability distribution $p$ for the initial Markov state takes the following values:
	\begin{align*}
		p(i)
		&=\begin{cases}
			\left(\frac{1}{2}\right)^{i+1}
				&0\leq i\leq m-2\\[0.5em]
			\left(\frac{1}{2}\right)^n
				&i=m-1
		\end{cases}
	\end{align*}
\end{observation}
The probability to have the first $i$ bits set to $1$ initially equals $2^{-i}$.
In case $i$ is less than $n=m-1$, the probability $\nicefrac{1}{2}$ to have a subsequent $0$ is multiplied to this term.
\begin{observation}\label{obs:M_leadingones}
	Let $m=n+1$ be the number of Markov states and $t=\nicefrac{\taumin}{\taumax}$ the pheromone ratio.
	The Markov matrix $M$ of dimension $(m\times m)$ is defined elementwise by $M(i,j)=\Pr[i\rightarrow j]$.
	The exact values are listed using the following partitioning of $M$:
	\begin{align*}
		\setlength\arraycolsep{2.5pt}
		\begin{pmatrix}
			(1) & (2) & \cdots & (2) & (3) \\[-1.4mm]
			0 & \ddots& \ddots & \vdots & \vdots \\[-1.7mm]
			\vdots & \ddots & \ddots & (2) & \vdots \\[-1.8mm]
			\vdots &  & \ddots & (1) & (3) \\[0.2mm]
			0 &  \cdots &  \cdots & 0 & (4) \\
		\end{pmatrix}
	\end{align*}
	\begin{enumerate}
		\item
			$0\leq i\leq m-2$:
			\begin{align*}
				M(i,i)
				=	1-\left(\frac{\taumax}{\taumax+\taumin}\right)^i\cdot\frac{\taumin}{\taumax+\taumin}
				=	1-t\cdot\left(\frac{1}{1+t}\right)^{i+1}
			\end{align*}
		\item
			$0\leq i<j\leq m-2$:
			\begin{align*}
				M(i,j)
				&=
					\left(\frac{\taumax}{\taumax+\taumin}\right)^i
					\cdot\frac{\taumin}{\taumax+\taumin}
					\cdot\left(\frac{1}{2}\right)^{j-i}\\
				&=
					t\cdot\left(\frac{1}{1+t}\right)^{i+1}\cdot\left(\frac{1}{2}\right)^{j-i}
			\end{align*}
		\item
			$0\leq i\leq m-2$:
			\begin{align*}
				M(i,m-1)
				&=
					\left(\frac{\taumax}{\taumax+\taumin}\right)^i
					\cdot\frac{\taumin}{\taumax+\taumin}
					\cdot\left(\frac{1}{2}\right)^{n-i-1}\\
				&=
					t\cdot\left(\frac{1}{1+t}\right)^{i+1}\cdot\left(\frac{1}{2}\right)^{n-i-1}
			\end{align*}
		\item
			$M(m-1,m-1)=0$
	\end{enumerate}
\end{observation}
In \baco, new candidate solutions that are worse than the current best solution are dismissed.
Thus the lower triangle of $M$ is $0$.
To calculate the remaining entries the following recurring structure is used.
An ant walking across $G$ has two options per bit.
One of them is weighted with $\taumax$ pheromones, the other one with $\taumin$.
Thus the probability to choose either of them is $\nicefrac{\taumax}{(\taumax+\taumin)}$ and $\nicefrac{\taumin}{(\taumax+\taumin)}$ respectively.
Diagonal entry $M(i,i)$, $0\le i\le m-2$, equals the probability to stay in state $i$.
This value is calculated via the counter-event where the current prefix of ones is extended by at least one bit.
Entry $M(m-1,m-1)$ is set to $0$ to represent the termination of the algorithm.
For an entry $M(i,j)$ of the upper triangle we differentiate between two cases.
If $j<m-1$ holds, then the ant needs to reproduce the path for the leading $i$ bits and afterwards pick the $\taumin$ options repeatedly until $j$ bits in total are set to $1$.
The subsequent bit needs to be set to $0$.
However, if $j=m-1$ holds, then that final $0$ does not exist.

Now the requirements for Theorem~\ref{thm:T_without_inverse} are fulfilled (straightforward calculation omitted) and it can be applied.
\begin{lemma}
	\label{lem:T_leadingones_without_inverse}
	Let $p$ and $M$ as in Observations~\ref{obs:p_leadingones} and~\ref{obs:M_leadingones}, respectively, and let $t>0$ be the pheromone ratio.
	The expected optimization time of \baco applied to \leadingones is the following:
	\begin{align}
		T(n,t)=\frac{1+t}{2t^2}\cdot\left((1+t)^n-1\right) \label{eq:expected_leadingones}
	\end{align}
\end{lemma}
To obtain this equation first calculate $\delta_j=(\nicefrac{1}{2t})\cdot(1+t)^{j+1}$, $j\le m-2$, from Eq.~\eqref{eq:delta_j} using the identity $A=\eye-M$ and cases (1) and (2) from Observation~\ref{obs:M_leadingones}.
Then use it for Theorem~\ref{thm:T_without_inverse} and simplify the term using the exact formula for partial geometric series.
Explicit bounds on the expected optimization time of \baco are deduced in the following theorem.
\begin{theorem}
	\label{thm:complexities_leadingones}
	Let $t>0$ be the pheronome ratio of \baco applied to \leadingones. Let furthermore $c$ be a positive constant.
	The expected optimization time $T(n,t)$ is bounded as listed in Table~\ref{tab:leadingones_runtimes}.
	\begin{proof}
		The pheromone ratio $t$ in Claim c1 is a positive constant $c$.
		Thus the constant prefactor $\frac{1+c}{2c^2}$ in Eq.~\eqref{eq:expected_leadingones} is omitted.
		Consequently, the bound on $T(n,t)$ is $\Theta((1+c)^n)$, i.\,e., $T(n,t)$ is exponentially growing.
		Start with Eq.~\eqref{eq:expected_leadingones} to prove Claim c2.
		Let $t=\frac{c}{n^s}$, $c>0$ and $s>0$.
		\begin{align*}
			\MoveEqLeft{T(n,t)
				\overset{\eqref{eq:expected_leadingones}}{=}
				\frac{1+\frac{c}{n^s}}{2\cdot\left(\frac{c}{n^s}\right)^2}\cdot\left(\left(1+\frac{c}{n^s}\right)^n-1\right)}\\
			\in{}&
				\Theta\left(n^{2s}\cdot\left(\left(1+\frac{c}{n^s}\right)^n-1\right)\right)\\
			={}&
				\Theta\left(n^{2s}\cdot\left(\left(\left(1+\frac{c}{n^s}\right)^{n^s}\right)^{\frac{n}{n^s}}-1\right)\right)\\
			={}&
				\Theta\left(n^{2s}\cdot\left(\left(\mathconst{e}^c\right)^{\frac{n}{n^s}}-1\right)\right)
			=
				\Theta\left(n^{2s}\cdot\left(\mathconst{e}^{c\cdot n^{1-s}}-1\right)\right)
		\end{align*}
		This proves Claim c2.
		For $c>0$ and $x\ge 0$ the claim
		\begin{align*}
			n^x\cdot\left(\mathconst{e}^{\frac{c}{n^x}}-1\right)
			\in\Theta(1)
		\end{align*}
		can be proven using L'Hospital's rule as follows:
		\begin{align*}
			\MoveEqLeft{\lim_{n\rightarrow\infty} n^x\cdot\left(\mathconst{e}^{\frac{c}{n^x}}-1\right)
			=
				\lim_{n\rightarrow\infty} \frac{\mathconst{e}^{\frac{c}{n^x}}-1}{n^{-x}}}\\
			\overset{\mathclap{\text{L'Ho.}}}{=}{}&
				\ \ \lim_{n\rightarrow\infty} \frac{-c\cdot x\cdot n^{-x-1}\cdot\mathconst{e}^{\frac{c}{n^x}}}{-x\cdot n^{-x-1}}
			=
				\lim_{n\rightarrow\infty} c\cdot\mathconst{e}^{\frac{c}{n^x}}
			\in
				\Theta(1)
		\end{align*}
		Apply this to the last term of the preceding calculation to prove Claim c3 with $s\ge 1$:
		\begin{align*}
			T(n,t)
			\in\Theta\left(n^{2s}\cdot\left(\mathconst{e}^{\frac{c}{n^{s-1}}}-1\right)\right)
			=\Theta\left(n^{2s-(s-1)}\right)
			=\Theta\left(n^{s+1}\right)
		\end{align*}
		To prove Claim c4, let $t\in\bigO(1)$ the pheromone ratio.
		Then the prefactor $1+t$ in Eq.~\eqref{eq:expected_leadingones} is bounded by $\Theta(1)$.
		Thus the remainder of Eq.~\eqref{eq:expected_leadingones} yields the stated bound.
	\end{proof}
\end{theorem}
\begin{table}
	\centering
	\caption{Expected optimization time of \baco applied to \leadingones depending on the pheromone ratio $t$ (Theorem~\ref{thm:complexities_leadingones}, Claims c1 through c4).}
	\label{tab:leadingones_runtimes}
	\begin{tabular}{lLL}
		\toprule
		&t=t(n)
		&T(n,t)\\
		\midrule
		c1.
		&c>0
		&\Theta\left((1+c)^n\right)\\[.5em]
		c2.
		&\frac{c}{n^s}, 0<s<1
		&\Theta\left(n^{2s}\cdot\left(\mathconst{e}^{c\cdot n^{1-s}}-1\right)\right)\\[.5em]
		c3.
		&\frac{c}{n^s}, s\ge 1
		&\Theta\left(n^{s+1}\right)\\[.5em]
		c4.
		&\mathcal{O}(1),t>0
		&\Theta\left(\frac{1}{t^2}\cdot\left((1+t)^n-1\right)\right)\\[.5em]
		\bottomrule
	\end{tabular}
\end{table}

%% file: sections/runtime_analysis/sorting.tex
\subsection{\textsc{Sorting}}
In this section, we analyze the runtime behavior of \baco for sorting a list of unique keys.
In contrast to the previous problem, \sorting is not a direct optimization problem and does not have a particular, obvious objective function.
In fact, there are
many ways to measure
the quality (``sortedness'') of a candidate solution such as the length of a longest ascending subsequence or the number of transpositions needed to sort the list of keys.
Here, the Final Position Prefix (\fpp) objective is introduced and used.
The \fpp objective maps a list of $n$ keys to its number of leading keys that are in the same position as in the sorted list.
This yields a number from $0$ to $n$, excluding $n-1$ since this case is impossible.
The evaluation of \fpp (inside the black box, not generating additional cost for the optimization time) has linear complexity regarding the number of comparisons of keys.
In fact, it takes roughly up to $n+\log_2 n$ comparisons when the following approach is used.
First identify the sorted prefix, i.\,e., the leading keys which are sorted in increasing order.
Then search for the minimum value in the suffix.
This minimum value determines which keys of the sorted prefix are actually in their final position, namely every key that is smaller.
This last step is realized via binary search.

The construction graph $G$ used for \sorting is a directed graph with $n+1$ vertices:
a starting node $v_{\text{start}}$ and nodes $v_0$ through $v_{n-1}$.
There is an edge from $v_{\text{start}}$ to every other node.
Nodes $v_0$ through $v_{n-1}$ induce a complete digraph.
Each node $v_i$, $0\le i\le n-1$, corresponds to one key.
The ant starts at $v_{\text{start}}$ and builds up a Hamiltonian path visiting every node exactly once.
The order of nodes yields an order of the keys that is evaluated with respect to
\fpp.

Let the Markov states be the numbers from $0$ through $n-1$.
States $0$ through $n-2$ are identified with the corresponding objective value of the current best solution.
The final state $n-1$ denotes the optimal \fpp value $n$.
\begin{observation}\label{obs:p_sorting}
	Let $m=n$ be the number of Markov states.
	The first path across $G$ constructed by the ant yields a uniformly at random chosen candidate solution for \sorting.
	The probability distribution vector $p$ for the initial Markov state takes the following values:
	\begin{align*}
		p(i)
		&=\begin{cases}
			(n-i-1)\cdot\frac{(n-i-1)!}{n!}
				&0\le i\le m-2\\
			\frac{1}{n!}
				&i=m-1
		\end{cases}
	\end{align*}
\end{observation}
The probability to initially have the first $i$ keys in their final position is $\nicefrac{1}{n}\cdot\nicefrac{1}{(n-1)}\cdots\nicefrac{1}{(n-i+1)}$.
If $i$ is smaller than $n$, then the probability $\nicefrac{(n-i-1)}{(n-i)}$ to have a subsequent key that is not in its final position is multiplied to this term.
\begin{observation}\label{obs:M_sorting}
	Let $m=n$ be the number of Markov states.
	The Markov matrix $M$ takes the following values analogously to Observation~\ref{obs:M_leadingones}:
	\begin{enumerate}
		\item
			$0\leq i\leq m-2$:
			$M(i,i)=1-t\cdot{\displaystyle\prod_{k=1}^{i+1}\frac{1}{1+(n-k)t}}$
		\item
			$0\leq i<j\leq m-2$:
			\begin{align*}
				M(i,j)
				=
				(n-j-1)\cdot t
				\cdot\left(\prod_{k=1}^{i+1}\frac{1}{1+(n-k)t}\right)
				\cdot\left(\prod_{k=i+1}^{j}\frac{1}{n-k}\right)
			\end{align*}
		\item
			$0\leq i\le m-2$:
			\begin{align*}
				M(i,m-1)
				=
				t
				\cdot\left(\prod_{k=1}^{i+1}\frac{1}{1+(n-k)t}\right)
				\cdot\left(\prod_{k=i+1}^{m-2}\frac{1}{n-k}\right)
			\end{align*}
		\item
			$M(m-1,m-1)=0$
	\end{enumerate}
\end{observation}
As mentioned in the Section~\ref{sec:leadingones}, matrix $M$ is of upper triangular shape since \baco discards solution candidates that are worse than the current best.
Entries $M(i,i)$ for $0\le i\le m-2$ are calculated using the counter-event in which the objective value is increased by at least $1$.
Starting at node $v_{\text{start}}$, the ant needs to take the $\taumax$-marked path for $i$ steps and then deviate to visit the correct successor that increases the \fpp objective value by one.
Assuming the ant has already visited $k$ nodes including $v_{\text{start}}$ and only $\taumax$-edges have been chosen, then the probability to again pick a $\taumax$-edge is
$\nicefrac{\taumax}{(\taumax+(n-k)\cdot\taumin)}=\nicefrac{1}{(1+(n-k)\cdot t)}$.
The formul\ae{} for the upper triangle start similarly, but the prefix of the current best solution is extended from $i$ to $j$ keys.
The pheromones do not appear in this part of the formul\ae{} since the suffix of a solution that is not part of the final position prefix has no influence on the \fpp objective and thus all permutations of that suffix are equally likely.
Analogously to \leadingones, if $j<n$ holds there needs to be a key after the final position prefix that is not in its final position and if $j$ equals $n$ no such key exists.

Now the requirements for Theorem~\ref{thm:T_without_inverse} are
satisfied (straightforward calculation omitted) and it can be applied.
\begin{lemma}
	\label{lem:T_sorting_without_inverse}
	Let $p$ and $M$ as in Observations~\ref{obs:p_sorting} and~\ref{obs:M_sorting}, respectively, and let $t>0$ be the pheromone ratio.
	The expected optimization time of \baco applied to \sorting is the following:
	\begin{align}
		T(n,t)
		={}&
			\frac{1}{t\cdot n!}
			\cdot\sum_{i=1}^{n-1}\Bigg(
				i\cdot i!
				\cdot\Bigg(
					\Bigg(\prod_{r=i}^{n-1} (1+r\cdot t)\Bigg)\nonumber\\
		&
					+\sum_{k=1}^{i-1}\Bigg(
						\frac{k}{k+1}
						\cdot\prod_{r=k}^{n-1} (1+r\cdot t)
					\Bigg)
				\Bigg)
			\Bigg) \label{eq:expected_sorting}
	\end{align}
\end{lemma}
To obtain this equation first calculate
\begin{align*}
	\delta_j=\frac{n-j-1}{n-j}\cdot\frac{1}{t}\cdot\prod_{k=1}^{j+1}(1+(n-k)\cdot t)
\end{align*}
from Eq.~\eqref{eq:delta_j} in the same way as for \leadingones.
Then use it for Theorem~\ref{thm:T_without_inverse} and simplify the term using index shifts.

We provide lower and upper bounds on the expected optimization time in order to enable further classification of the underlying complexity.
\begin{lemma}
	\label{lem:sorting_bounds_small_t}
	For any pheromone ratio $t>0$ the expected optimization time of \baco applied to \sorting is bounded as follows:
	\begin{equation}
		\frac{n-2}{2t}\le T(n,t)\le\frac{n}{t}\cdot(1+n\cdot t)^n \label{eq:sorting_bounds_small_t}
	\end{equation}
	\begin{proof}
		\allowdisplaybreaks
		The following identity can easily be proven for all $n\in\mathbb{N}$ by mathematical induction:
		\begin{align}
			\sum_{i=1}^{n-1} i^2\cdot i!
			=(n-1)\cdot n!-\sum_{i=1}^{n-1} i!
			\label{eq:square_weighted_factorial_sum}
		\end{align}
		To prove the lower bound on $T(n,t)$ start with its definition:
		\begin{align*}
			T(n,t)
			\overset{\eqref{eq:expected_sorting}}{=}{}&
				\frac{1}{t\cdot n!}
				\cdot\sum_{i=1}^{n-1}\Bigg(
					i\cdot i!
					\cdot\Bigg(
						\Bigg(\prod_{r=i}^{n-1} \underbrace{(1+r\cdot t)}_{\ge 1}\Bigg)\\
			&
						+\sum_{k=1}^{i-1}\Bigg(
							\underbrace{\frac{k}{k+1}}_{\ge\frac{1}{2}}
							\cdot\prod_{r=k}^{n-1} \underbrace{(1+r\cdot t)}_{\ge 1}
						\Bigg)
					\Bigg)
				\Bigg)\\
			\ge{}&
				\frac{1}{t\cdot n!}
				\cdot\sum_{i=1}^{n-1}\Bigg(
					i\cdot i!
					\cdot\Bigg(
						1+\frac{1}{2}\cdot\sum_{k=1}^{i-1} 1
					\Bigg)
				\Bigg)
			\ge
				\frac{1}{2t\cdot n!}
				\cdot\sum_{i=1}^{n-1} \left(i^2\cdot i!\right)\\
			\overset{\eqref{eq:square_weighted_factorial_sum}}{=}{}&
				\frac{1}{2t\cdot n!}
				\cdot\left(
					(n-1)\cdot n!-\sum_{i=1}^{n-1} i!
				\right)
			=
				\frac{1}{2t}
				\cdot\Bigg(
					n-1-\sum_{i=1}^{n-1} \frac{i!}{n!}
				\Bigg)\\
			\ge{}&
				\frac{n-2}{2t}
		\end{align*}
		The upper bound is derived as follows:
		\begin{align*}
			T(n,t)
			\overset{\eqref{eq:expected_sorting}}{=}{}&
				\frac{1}{t\cdot n!}
				\cdot\sum_{i=1}^{n-1}\Bigg(
					i\cdot i!
					\cdot\Bigg(
						\Bigg(\prod_{r=i}^{n-1} (1+r\cdot t)\Bigg)\\
			&
						+\sum_{k=1}^{i-1}\Bigg(
							\frac{k}{k+1}
							\cdot\prod_{r=k}^{n-1} (1+r\cdot t)
						\Bigg)
					\Bigg)
				\Bigg)\\
			\le{}&
				\frac{1}{t\cdot n!}
				\cdot\sum_{i=1}^{n-1}\Bigg(
					i\cdot i!
					\cdot\Bigg(
						(1+n\cdot t)^n
						+\sum_{k=1}^{i-1} (1+n\cdot t)^n
					\Bigg)
				\Bigg)\\
			={}&
				\frac{1}{t\cdot n!}
				\cdot\sum_{i=1}^{n-1}\Bigg(
					i^2\cdot i!\cdot (1+n\cdot t)^n
				\Bigg)\\
			\overset{\eqref{eq:square_weighted_factorial_sum}}{=}{}&
				\frac{(1+n\cdot t)^n}{t\cdot n!}
				\cdot\left(
					(n-1)\cdot n!-\sum_{i=1}^{n-1} i!
				\right)
			\le
				\frac{n}{t}\cdot(1+n\cdot t)^n\qedhere
		\end{align*}
	\end{proof}
\end{lemma}
Using these bounds on the expected optimization time we further classify the underlying complexity of \baco for \sorting with respect to various pheromone ratios $t$.
\begin{theorem}
	\label{thm:complexities_sorting}
	Let $t$ be the pheronome ratio of \baco applied to \sorting.
	The expected optimization time $T(n,t)$ is bounded as listed in Table~\ref{tab:sorting_runtimes}.
	\begin{proof}
		Use Lemma~\ref{lem:T_sorting_without_inverse} to prove Claim c5:
		\begin{align*}
			T(n,t)
			\overset{\eqref{eq:expected_sorting}}{=}{}&
			\frac{1}{t\cdot n!}
				\cdot\sum_{i=1}^{n-1}\Bigg(
					i\cdot i!
					\cdot\Bigg(
						\Bigg(\prod_{r=i}^{n-1} (1+r\cdot t)\Bigg)\\
			&
						+\sum_{k=1}^{i-1}\Bigg(
							\frac{k}{k+1}
							\cdot\prod_{r=k}^{n-1} (1+r\cdot t)
						\Bigg)
					\Bigg)
				\Bigg)\\
			\ge{}&
				\frac{1}{2\cdot t\cdot n!}
				\cdot\sum_{i=1}^{n-1}\Bigg(
					i\cdot i!
					\cdot\Bigg(
						\Bigg(\sum_{r=i}^{n-1} r\cdot t\Bigg)
						+\sum_{k=1}^{i-1}\sum_{r=k}^{n-1} r\cdot t
					\Bigg)
				\Bigg)\\
			={}&
				\frac{1}{2\cdot n!}
				\cdot\sum_{i=1}^{n-1}\Bigg(
					i\cdot i!
					\cdot\sum_{k=1}^{i}\sum_{r=k}^{n-1} r
				\Bigg)
		\end{align*}
		Calculate the double sum explicitly and apply Eq.~\eqref{eq:square_weighted_factorial_sum} to prove that $T(n,t)\in\Omega\left(n^3\right)$ holds.
		Next prove Claim c6.
		Based on Lemma~\ref{lem:sorting_bounds_small_t}, the lower bound $\Omega\left(\frac{n}{t}\right)$ is obvious while the upper bound is derived as follows.
		Let $t\in\mathcal{O}\big(\frac{1}{n^2}\big)$.
		Then $n^2\cdot t\in\mathcal{O}(1)$ and $\exp(n^2\cdot t)\in\mathcal{O}(1)$ hold.
		Use this in the following calculation.
		\begin{align*}
			T(n,t)
			\overset{\eqref{eq:sorting_bounds_small_t}}{\le}
				\frac{n}{t}\cdot(1+n\cdot t)^n
			=
				\frac{n}{t}\cdot\left(1+\frac{n^2\cdot t}{n}\right)^n
			\le
				\frac{n}{t}\cdot\mathconst{e}^{n^2\cdot t}
			\in
				\mathcal{O}\left(\frac{n}{t}\right)
		\end{align*}
		Claim c7 is a special case of c6.
		To prove Claim c8, use the upper bound in Lemma~\ref{lem:sorting_bounds_small_t} and set $t=\nicefrac{c}{n^s},c>0,s<2$ for the following calculation:
		\begin{align*}
			\MoveEqLeft{T(n,t)
			\overset{\eqref{eq:sorting_bounds_small_t}}{\le}
				\frac{n}{\frac{c}{n^s}}\cdot\left(1+n\cdot \frac{c}{n^s}\right)^n
			=
				\frac{n^{s+1}}{c}\cdot\left(1+\frac{c\cdot n^{2-s}}{n}\right)^n}\\
			\le{}&
				\frac{n^{s+1}}{c}\cdot\mathconst{e}^{c\cdot n^{2-s}}
			\in
				\mathcal{O}\left(n^{s+1}\cdot\mathconst{e}^{c\cdot n^{2-s}}\right)
		\end{align*}
		In Claim c9, $t$ is set to $1$ which results in a blind search.
		For the proof, the following identity is utilized:
		\begin{align}
			\forall n\in\mathbb{N}:\quad\sum_{i=1}^{n-1} (i\cdot i!)=n!-1
			\label{eq:linear_weighted_factorial_sum}
		\end{align}
		With that the lower bound is calculated.
		\begin{align*}
			T(n,t)
			\overset{\eqref{eq:expected_sorting}}{=}{}&
				\frac{1}{n!}
				\cdot\sum_{i=1}^{n-1}\Bigg(
					i\cdot i!
					\cdot\Bigg(
						\Bigg(\prod_{r=i}^{n-1} (1+r)\Bigg)\\
			&
						+\sum_{k=1}^{i-1}\Bigg(
							\frac{k}{k+1}
							\cdot\prod_{r=k}^{n-1} (1+r)
						\Bigg)
					\Bigg)
				\Bigg)\\
			={}&
				\frac{1}{n!}
				\cdot\sum_{i=1}^{n-1}\Bigg(
					i\cdot i!
					\cdot\Bigg(
						\frac{n!}{i!}
						+\sum_{k=1}^{i-1}\Bigg(
							\frac{k}{k+1}
							\cdot\frac{n!}{k!}
						\Bigg)
					\Bigg)
				\Bigg)\\
			\ge{}&
				\frac{1}{n!}
				\cdot\sum_{i=1}^{n-1}\Bigg(
					i\cdot i!\cdot\frac{1}{2}
					\cdot\sum_{k=1}^{i} \frac{n!}{k!}
				\Bigg)
			=
				\frac{1}{2}
				\cdot\sum_{i=1}^{n-1}\Bigg(
					i\cdot i!
					\cdot\sum_{k=1}^{i} \frac{1}{k!}
				\Bigg)\\
			\ge{}&
				\frac{1}{2}
				\cdot\sum_{i=1}^{n-1} (i\cdot i!)
			\overset{\eqref{eq:linear_weighted_factorial_sum}}{=}
				\frac{1}{2}\cdot(n!-1)
			\in
				\Omega(n!)
		\end{align*}
		The proof of the upper bound starts analogously (left out, cf.~above):
		\begin{align*}
			\MoveEqLeft{T(n,t)
			\overset{\eqref{eq:expected_sorting}}{=}
				\dots
			\le
				\frac{1}{n!}
				\cdot\sum_{i=1}^{n-1}\left(
					i\cdot i!
					\cdot\sum_{k=1}^{i} \frac{n!}{k!}
				\right)
			\le
				\sum_{i=1}^{n-1} \left(i\cdot i!\cdot\sum_{k=1}^{\infty} \frac{1}{k!}\right)}\\
			\overset{\eqref{eq:linear_weighted_factorial_sum}}{=}{}&
				(n!-1)\cdot (e-1)
			\in
				\mathcal{O}(n!)
				\qedhere
		\end{align*}
	\end{proof}
\end{theorem}
\begin{table}
	\centering
	\caption{Expected optimization time of \baco applied to \sorting depending on the pheromone ratio $t$ (Theorem~\ref{thm:complexities_sorting}, Claims c5 through c9).}
	\label{tab:sorting_runtimes}
	\begin{tabular}{lLL}
		\toprule
		&t=t(n)
		&T(n,t)\\
		\midrule
		c5.
		&t(n)>0
		&\Omega\left(n^3\right)\\
		c6.
		&\mathcal{O}\left(\frac{1}{n^2}\right)
		&\Theta\left(\frac{n}{t}\right)\\
		c7.
		&\frac{c}{n^s},s\ge 2
		&\Theta\left(n^{s+1}\right)\\
		c8.
		&\frac{c}{n^s},s<2
		&\mathcal{O}\left(n^{s+1}\cdot\mathconst{e}^{c\cdot n^{2-s}}\right)\\
		c9.
		&1
		&\Theta\left(n!\right)\\
		\bottomrule
	\end{tabular}
\end{table}

%% file: sections/runtime_analysis/table_runtime_comparison.tex
\begin{table}
	\centering
	\caption{Expected optimization time bounds of \baco, \oneoneea \cite{DROSTE200251} and \onepso \cite{Muehlenthaler2021} applied to \leadingones and \sorting.}
	\label{tab:comparison_runtime_aco_ea_pso}
	\begin{tabular}{lLL}
		\toprule
		&\leadingones
		&\sorting
		\\
		\midrule
		\baco
			&\Theta\left(n^2\right)
			&\Theta\left(n^3\right)
		\\[.3em]
		\oneoneea
			&\Theta\left(n^2\right)
			&\Theta\left(n^2\cdot\log n\right)
		\\[.3em]
		\onepso
			&\text{--}
			&\Omega\left(n^2\right),\mathcal{O}\left(n^2\cdot\log n\right)
		\\
		\bottomrule
	\end{tabular}
\end{table}

%% file: sections/experiments/experiments.tex
\section{Experiments}
\label{sec:experiments}
\pgfplotsset{
	layers/my layer set/.define layer set={
		background,
		main,
		foreground
	}{
	},
	set layers=my layer set,
}
In this section the theoretical results are validated.
For this, an implementation of \baco written in C++ is used to solve \leadingones and \sorting for a range of problem sizes $n$.
Figure~\ref{fig:experimental_optimization_time_leadingones_1_n} visualizes the optimization time of the implementation applied to \leadingones.
We ran $20$ repetitions for each problem size $n$ from $5$ up to~$200$.
The analytically calculated expected optimization time $T(n,t)$ for \leadingones from Lemma~\ref{lem:T_leadingones_without_inverse} is plotted in red.
Both, the implementation and the exact formula used the optimal pheromone ratio $t=\nicefrac{1}{n}$.
Analogously, Figure~\ref{fig:experimental_optimization_time_sorting_1_nn} shows the optimization times for \sorting.
We ran $40$ repetitions for each problem size $n$ from $5$ up to $100$.
The analytically calculated expected optimization time $T(n,t)$ for \sorting from Lemma~\ref{lem:T_sorting_without_inverse} is plotted in red.
For this experiment the implementation and the exact formula used the optimal pheromone ratio $t=\nicefrac{1}{n^2}$.
As can clearly be seen in both plots, the experimental results fit the theoretical findings.
\begin{figure}
	\centering
	
  \input{sections/experiments/T_leadingones_1_n.tikz}%

	\caption{
		Optimization time for \leadingones and the analytically calculated expected optimization time using the pheromone ratio $t=\frac{1}{n}$.
	}
	\label{fig:experimental_optimization_time_leadingones_1_n}
\end{figure}
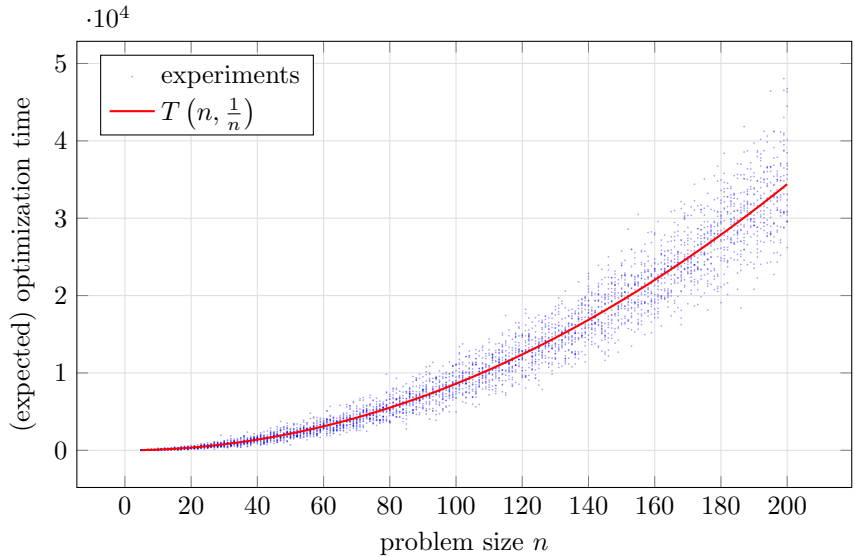
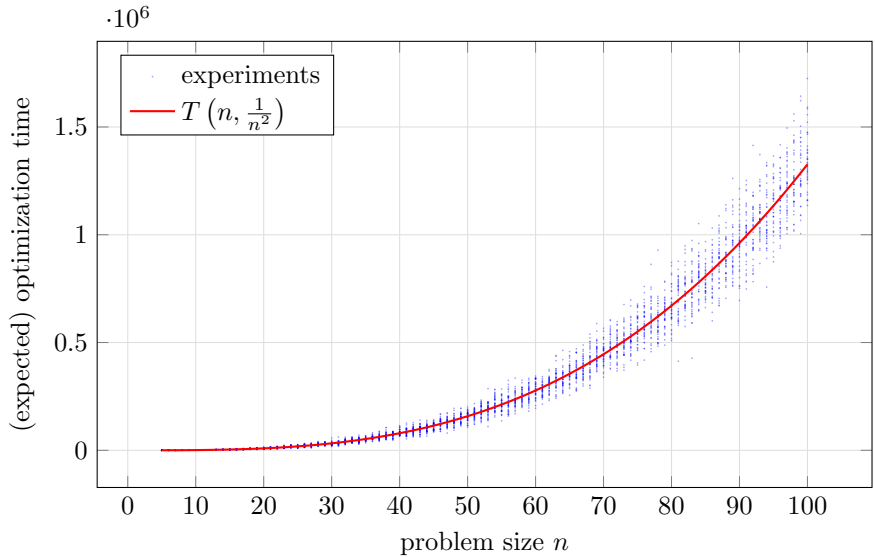
\begin{figure}
	\centering
	
  \input{sections/experiments/T_sorting_1_nn.tikz}%

	\caption{
		Optimization time for \sorting and the analytically calculated expected optimization time using the pheromone ratio $t=\frac{1}{n^2}$.
	}
	\label{fig:experimental_optimization_time_sorting_1_nn}
\end{figure}

%% file: sections/experiments/T_leadingones_1_n.tikz
\begin{tikzpicture}
	\begin{axis}[
		xlabel={problem size $n$},
		ylabel={(expected) optimization time},
		grid=both,
		minor grid style={gray!25},
		major grid style={gray!25},
		width=0.98\linewidth,
		height=0.62\linewidth,
		no marks,
		legend cell align=left,
		legend pos=north west,
	]
		\addplot[
			only marks,
			mark size=0.1pt,
			blue,
			opacity=0.4,
		]
		table[
			x=n,
			y=iterations,
			col sep=comma
		]{sections/experiments/data/leadingones_1_n_n_5_1_200.csv};
		\addlegendentry{experiments};
		\addplot[
			red,
			on layer=foreground,
			thick,
		]
		table[
			x=n,
			y=T,
			col sep=comma
		]{sections/experiments/data/leadingones_1_n_T_and_bounds_n_5_1_200.csv};
		\addlegendentry{$T\left(n,\frac{1}{n}\right)$};
	\end{axis}
\end{tikzpicture}

%% file: sections/experiments/T_sorting_1_nn.tikz
\begin{tikzpicture}
	\begin{axis}[
		xlabel={problem size $n$},
		ylabel={(expected) optimization time},
		grid=both,
		minor grid style={gray!25},
		major grid style={gray!25},
		width=0.98\linewidth,
		height=0.62\linewidth,
		no marks,
		legend cell align=left,
		legend pos=north west,
	]
		\addplot[
			only marks,
			mark size=0.1pt,
			blue, 
			opacity=0.4,
		]
		table[
			x=n,
			y=iterations,
			col sep=comma
		]{sections/experiments/data/sorting_1_nn_n_5_1_100.csv};
		\addlegendentry{experiments}
		\addplot[
			red,
			on layer=foreground,
			thick,
		]
		table[
			x=n,
			y=T_ex,
			col sep=comma
		]{sections/experiments/data/sorting_1_nn_T_and_bounds_5_1_100.csv};
		\addlegendentry{$T\left(n,\frac{1}{n^2}\right)$}
	\end{axis}
\end{tikzpicture}

%% file: sections/conclusion.tex
\section{Conclusion \& Outlook}
In this work, the ACO variant Bivalent Ant Colony Optimization (\baco) was investigated and analyzed based on the problems \sorting and \leadingones.
The main characteristic of \baco is the usage of bivalent pheromones, i.\,e., pheromones can only take the values $\taumin$ or $\taumax$.
We provided a Markov chain-based approach that allows to deduce an exact formula for the expected optimization time of \baco and demonstrate it using both problems.
The influence of the pheromone ratio $t=\nicefrac{\taumin}{\taumax}$ is presented precisely.
With the help of these formul\ae{} we prove the expected optimization time complexities $\Theta(n^3)$ for \sorting and $\Theta(n^2)$ for \leadingones that are achieved using the optimal values for $t$, respectively.
To the best of our knowledge, bounds for \sorting and the lower bound for \leadingones have been unknown so far.
In the domain of \sorting \baco is still outperformed by the \oneoneea and \onepso by a factor of $\nicefrac{n}{\log n}$.
However, our analysis together with a different objective function is possibly the right approach to achieve a draw between those three algorithms.
Additionally, the missing lower bound for \leadingones yields the identical complexity $\Theta(n^2)$ as the \oneoneea.
Experiments with an implementation of the algorithm validated the results proven by theoretical analysis.

%% file: sections/appendix/appendix.tex
\section{Infinite sum of matrix powers}
\label{sec:infinite_sum_of_matrix_powers}
Let $M$ be the Markov matrix from Section~\ref{sec:runtime_analysis}.
Then $M$ is an upper triangular matrix and all entries on the diagonal are strictly smaller than $1$ and non-negative.
This implies that $(\eye-M)$ is also an upper triangular matrix and every diagonal entry is greater than $0$.
Thus the inverse matrix of $(\eye-M)$ exists and the following identity is well-defined and will be proven afterwards:
$$ 
	\sum_{i=0}^{\infty} i\cdot M^i
	=(\eye-M)^{-1}\cdot(\eye-M)^{-1}\cdot M
$$
\begin{proof}
	\begin{align*}
		\MoveEqLeft{(\eye-M)\cdot(\eye-M)\cdot\sum_{i=0}^{\infty} i\cdot M^i
		=	\left(\eye-2\cdot M+M^2\right)\cdot\sum_{i=1}^{\infty} i\cdot M^i}\\
		={}&
				\left(\sum_{i=1}^{\infty}  i\cdot M^i    \right)
		-		\left(\sum_{i=1}^{\infty} 2i\cdot M^{i+1}\right)
		+		\left(\sum_{i=1}^{\infty}  i\cdot M^{i+2}\right)\\
		={}&
				\left(\sum_{i=1}^{\infty}   i   \cdot M^i\right)
		-		\left(\sum_{i=2}^{\infty} 2(i-1)\cdot M^i\right)
		+		\left(\sum_{i=3}^{\infty}  (i-2)\cdot M^i\right)\\
		={}&
				M
		+		\sum_{i=3}^{\infty} (i-2(i-1)+(i-2))\cdot M^i
		=		M
	\end{align*}
	Multiplication by $(\eye-M)^{-1}\cdot(\eye-M)^{-1}$ from the left yields the statement.
	Furthermore the infinite sum of powers of $M$ can be reordered since it converges as the eigenvalues of $M$ are between $0$ and $1$.
\end{proof}


%% file: arxiv.bbl
\begin{thebibliography}{10}

\bibitem{abounacer:hal-01286691}
{\sc Abounacer, R., Boukachour, J., Dkhissi, B., and El~Hilali~Alaoui, A.}
\newblock {A hybrid {Ant Colony Algorithm} for the exam timetabling problem}.
\newblock {\em {Revue Africaine de Recherche en Informatique et
  Math{\'e}matiques Appliqu{\'e}es} 12\/} (2010), 15--42.

\bibitem{4424704}
{\sc Doerr, B., Happ, E., and Klein, C.}
\newblock A tight analysis of the (1 + 1)-ea for the single source shortest
  path problem.
\newblock In {\em 2007 IEEE Congress on Evolutionary Computation\/} (2007),
  pp.~1890--1895.

\bibitem{doerr2020}
{\sc Doerr, B., and Neumann, F.}, Eds.
\newblock {\em Theory of Evolutionary Computation---Recent Developments in
  Discrete Optimization}.
\newblock Springer, 2020.
\newblock Also available at
  \url{http://www.lix.polytechnique.fr/Labo/Benjamin.Doerr/doerr_neumann_book.html}.

\bibitem{doerr2007a}
{\sc Doerr, B., Neumann, F., Sudholt, D., and Witt, C.}
\newblock On the runtime analysis of the 1-ant aco algorithm.
\newblock In {\em Proceedings of the 9th Annual Conference on Genetic and
  Evolutionary Computation\/} (New York, NY, USA, 2007), GECCO '07, Association
  for Computing Machinery, pp.~33--40.

\bibitem{DOERR20111629}
{\sc Doerr, B., Neumann, F., Sudholt, D., and Witt, C.}
\newblock Runtime analysis of the 1-ant ant colony optimizer.
\newblock {\em Theoretical Computer Science 412}, 17 (2011), 1629--1644.

\bibitem{dorigo1992}
{\sc Dorigo, M.}
\newblock {\em Optimization, Learning and Natural Algorithms}.
\newblock PhD thesis, Dipartimento di Elettronica, Politecnico di Milano,
  Italy, 1992.
\newblock In Italian.

\bibitem{DORIGO2005243}
{\sc Dorigo, M., and Blum, C.}
\newblock Ant colony optimization theory: A survey.
\newblock {\em Theoretical Computer Science 344}, 2 (2005), 243--278.

\bibitem{782657}
{\sc Dorigo, M., and Di~Caro, G.}
\newblock Ant colony optimization: a new meta-heuristic.
\newblock In {\em Proc 1999 Congress on Evolutionary Computation (CEC)\/}
  (Washington, DC, USA, 1999), vol.~2, IEEE, pp.~1470--1477.

\bibitem{dorigo1999a}
{\sc Dorigo, M., Maniezzo, V., and Colorni, A.}
\newblock Positive feedback as a search strategy.
\newblock {\em Tech rep., 91-016, Dip Elettronica, Politecnico di Milano,
  Italy\/} (04 1999).

\bibitem{DROSTE200251}
{\sc Droste, S., Jansen, T., and Wegener, I.}
\newblock On the analysis of the (1+1) evolutionary algorithm.
\newblock {\em Theoretical Computer Science 276}, 1 (2002), 51--81.

\bibitem{gambardella1996}
{\sc Gambardella, L., and Dorigo, M.}
\newblock Solving symmetric and asymmetric tsps by ant colonies.
\newblock In {\em Proceedings of IEEE International Conference on Evolutionary
  Computation\/} (1996), pp.~622--627.

\bibitem{GUTJAHR20082711}
{\sc Gutjahr, W.~J.}
\newblock First steps to the runtime complexity analysis of ant colony
  optimization.
\newblock {\em Computers \& Operations Research 35}, 9 (2008), 2711--2727.
\newblock Part of Special Issue: Bio-inspired Methods in Combinatorial
  Optimization.

\bibitem{koetzing2010}
{\sc K\"{o}tzing, T., Lehre, P.~K., Neumann, F., and Oliveto, P.~S.}
\newblock Ant colony optimization and the minimum cut problem.
\newblock In {\em Proceedings of the 12th Annual Conference on Genetic and
  Evolutionary Computation\/} (New York, NY, USA, 2010), GECCO '10, Association
  for Computing Machinery, pp.~1393--1400.

\bibitem{Muehlenthaler2021}
{\sc M{\"u}hlenthaler, M., Ra{\ss}, A., Schmitt, M., and Wanka, R.}
\newblock Exact {M}arkov chain-based runtime analysis of a discrete particle
  swarm optimization algorithm on sorting and {OneMax}.
\newblock {\em Natural Computing 21\/} (2021), 651--677.

\bibitem{neumann2006}
{\sc Neumann, F., and Witt, C.}
\newblock Runtime analysis of a simple ant colony optimization algorithm.
\newblock In {\em Proceedings of the 17th International Conference on
  Algorithms and Computation\/} (Berlin, Heidelberg, 2006), ISAAC'06,
  Springer-Verlag, pp.~618--627.

\bibitem{neumann2009runtime}
{\sc Neumann, F., and Witt, C.}
\newblock Runtime analysis of a simple ant colony optimization algorithm.
\newblock {\em Algorithmica 54}, 2 (2009), 243--255.

\bibitem{NEUMANN20102406}
{\sc Neumann, F., and Witt, C.}
\newblock Ant colony optimization and the minimum spanning tree problem.
\newblock {\em Theoretical Computer Science 411}, 25 (2010), 2406--2413.

\bibitem{ScharnowTW04}
{\sc Scharnow, J., Tinnefeld, K., and Wegener, I.}
\newblock The analysis of evolutionary algorithms on sorting and shortest paths
  problems.
\newblock {\em J. Math. Model. Algorithms 3}, 4 (2004), 349--366.

\bibitem{Shyu2004}
{\sc Shyu, S.~J., Yin, P.-Y., and Lin, B.~M.}
\newblock An {Ant Colony Optimization} algorithm for the minimum weight vertex
  cover problem.
\newblock {\em Annals of Operations Research 13}, 1 (2004), 283--304.

\bibitem{stutzle1997max}
{\sc St{\"u}tzle, T., and Hoos, H.}
\newblock Max-min ant system and local search for the traveling salesman
  problem.
\newblock In {\em Proceedings of 1997 IEEE international conference on
  evolutionary computation (ICEC'97)\/} (1997), IEEE, pp.~309--314.

\bibitem{STUTZLE2000889}
{\sc St{\"u}tzle, T., and Hoos, H.~H.}
\newblock Max-min ant system.
\newblock {\em Future Generation Computer Systems 16}, 8 (2000), 889--914.

\bibitem{SUDHOLT2012165}
{\sc Sudholt, D., and Thyssen, C.}
\newblock Running time analysis of {Ant Colony Optimization} for shortest path
  problems.
\newblock {\em Journal of Discrete Algorithms 10\/} (2012), 165--180.

\end{thebibliography}
